\documentclass[10pt,twocolumn,letterpaper]{article}

\usepackage{iccv}
\usepackage{times}
\usepackage{epsfig}
\usepackage{graphicx}
\usepackage{amsmath}
\usepackage{amssymb}

\usepackage{multirow}
\usepackage{booktabs}
\usepackage{setspace}
\usepackage{color}
\usepackage{algorithm}
\usepackage{algpseudocode}
\usepackage{algorithmicx}

\usepackage{stfloats}
\usepackage{threeparttable}
\usepackage{stfloats}

\newcommand{\tabincell}[2]{\begin{tabular}{@{}#1@{}}#2\end{tabular}}
\usepackage[square,sort,sort&compress,numbers]{natbib}

\usepackage[pagebackref=true,breaklinks=true,letterpaper=true,colorlinks,bookmarks=false]{hyperref}
\iccvfinalcopy 

\def\iccvPaperID{6497} 

\ificcvfinal\fi

\begin{document}

\title{Group-wise Inhibition based Feature Regularization for Robust Classification}

\author{Haozhe Liu$^\dagger$, Haoqian Wu$^\dagger$, Weicheng Xie$^*$,  Feng Liu$^*$, Linlin Shen\\
$^1$ Computer Vision Institute, College of Computer Science and Software Engineering, \\
$^2$ SZU Branch, Shenzhen Institute of Artificial Intelligence and Robotics for Society \\
$^3$ National Engineering Laboratory for Big Data System Computing Technology \\
$^4$Guangdong Key Laboratory of Intelligent Information Processing, \\
Shenzhen University, Shenzhen 518060, China\\
{\tt\small \{liuhaozhe2019, wuhaoqian2019\}@email.szu.edu.cn,}
{\tt\small \{wcxie,feng.liu,llshen\}@szu.edu.cn}
}

\maketitle
\ificcvfinal\thispagestyle{empty}\fi

\begin{abstract}
The convolutional neural network (CNN) is vulnerable to degraded images with even very small variations (e.g. corrupted and adversarial samples).
One of the possible reasons is that CNN pays more attention to the most discriminative regions, but ignores the auxiliary features when learning, leading to the lack of feature diversity for final judgment.
In our method, we propose to dynamically suppress significant activation values of CNN  by group-wise inhibition, but not fixedly or randomly handle them when training. The feature maps with different activation distribution are then processed separately to take the feature independence into account. CNN is finally guided to learn richer discriminative features hierarchically for robust classification according to the proposed regularization.
Our method is comprehensively evaluated under multiple settings, including classification against corruptions, adversarial attacks and low data regime.
Extensive experimental results show that the proposed method can achieve significant improvements in terms of both robustness and generalization performances, when compared with the state-of-the-art methods. Code is available at \url{https://github.com/LinusWu/TENET_Training}.
\end{abstract}
\footnotetext{$^\dagger$Equal Contribution}
\footnotetext{$^*$Corresponding Author}
\section{Introduction}
\label{sec:Intro}
Recent advances in convolutional neural networks (CNNs) have led to far-reaching improvements in computer vision tasks \cite{he2016deep,lecun2015deep}.
However, vulnerability of CNNs to image variations, including image corruptions \cite{goodfellow2014explaining} and adversarial samples \cite{dong2020benchmarking}, has not been well resolved yet. Researchers are thus exploring various ways to improve the network robustness against these variations.

\begin{figure}
  \centering
  \includegraphics[width=.47\textwidth]{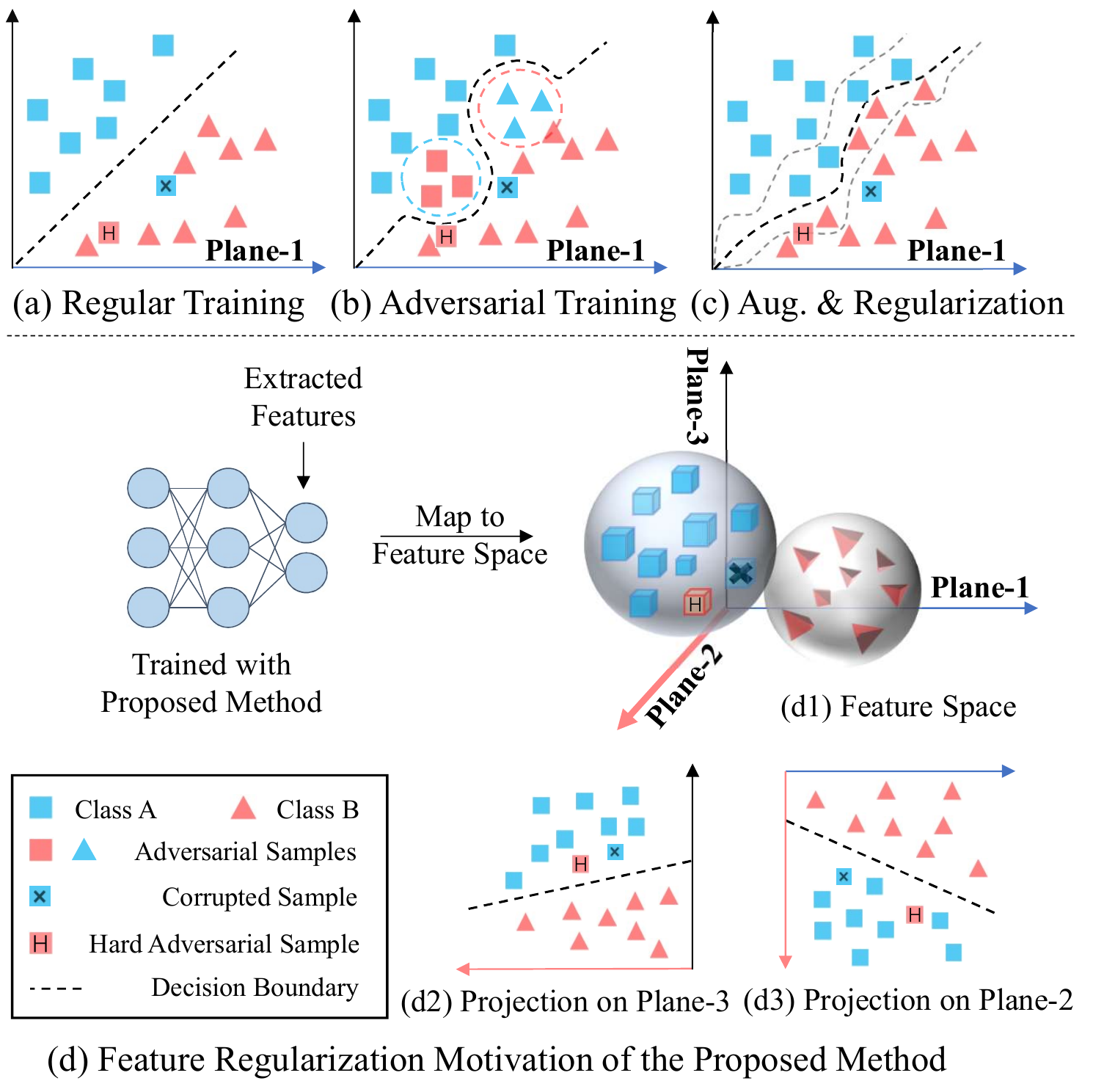}
  \caption{Some solutions to improve the robustness of CNN. Unlike with the regular training (a), adversarial training (b) widely utilizes adversarial samples to train a robust CNN. Data augmentation and regularization based method (c) improves the robustness performance by filling up new samples surrounding the decision boundary. The proposed regularization method (d) enables network to increase the representation space (e.g. red auxiliary axis in \textit{d1}) of the features learned by the CNN,
  and achieves better robustness against corrupted and adversarial samples, with various projections on new planes (e.g. \textit{d2} and \textit{d3}). Best viewed in color.
  }
  \label{fig:task}
\end{figure}
\begin{figure}
  \centering
  \includegraphics[width=.38\textwidth]{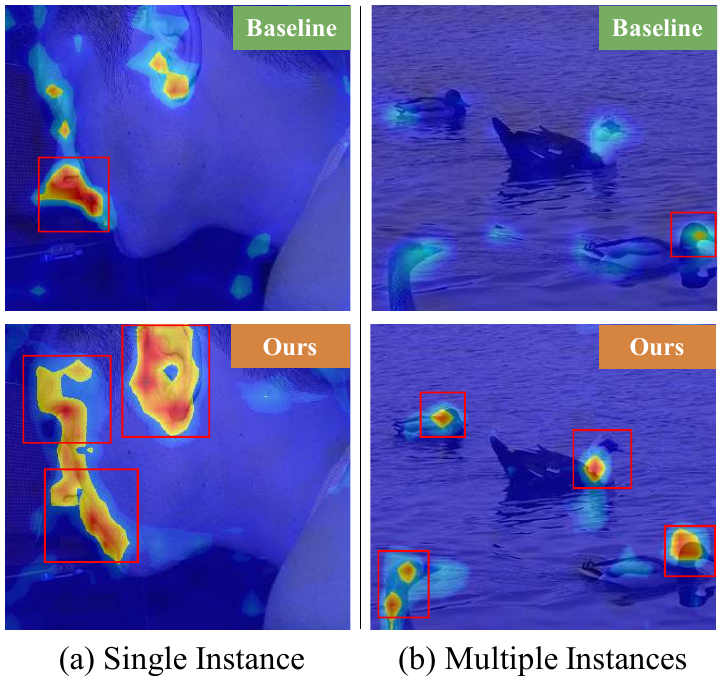}
  \caption{The heatmap visualization of feature maps encoded with ResNet-50, based on Grad-CAM \cite{zhou2016learning,selvaraju2017grad} with or without the proposed method.
  Our method locates more diverse discriminative regions (in red boxes) for both single-instance (a) and multiple-instance  (b) samples.
  }
  \label{fig:problem}
\end{figure}

Adversarial training \cite{goodfellow2014explaining,zheng2020efficient,xie2019feature} is a typical solution to improve the robustness of CNNs, which includes the attacked samples into the  training data, as shown in Fig. \ref{fig:task} (b).
Since adversarial training may impair the generalization performance,
there is often an inherent trade-off between classification accuracy and adversarial robustness \cite{xie2019feature, xie2020adversarial}.
In order to improve the robustness and generalization simultaneously, data augmentation and regularization methods (e.g. Random Erasing \cite{zhong2020random}, Augmix \cite{hendrycks2019augmix}, Cutout\cite{devries2017improved}, Dropout \cite{hinton2012improving} and DeepAugmentation\cite{hendrycks2020many}) are proposed. As shown in Fig. \ref{fig:task} (c), these algorithms address data augmentation by randomly generating new samples obeying the same distribution as the training data. Generally, data regularization methods are state-agnostic, which can not be dynamically adjusted during CNN training.
Thus, these regularization techniques of CNNs
\cite{choe2019attention,wang2020self,hou2018self,wei2017object} failed to learn features with  sufficient diversity.
As shown in the first row of Fig. \ref{fig:problem}, CNNs can locate the most discriminative regions \cite{zhou2016learning} for both  single-instance and multi-instance samples with the regularization method, while neglecting other auxiliary features that are critical for the recognition.
The lack of auxiliary features may lead to insufficient feature diversity, which consequently results in a feature space
with low-dimension for classification and limits the robustness.
Meanwhile, current adversarial training and regularization methods concentrate on the global  image information by expanding the training set, while the independence of local features is not fully explored.
These limitations motivate us to improve the diversity of extracted features by CNNs and devise a non-image-wise regularization strategy to enhance network robustness.

In this paper, we propose a group-wise inhibition based regularization method for improving feature diversity and network robustness, denoted as TENET Training.
Fig.\ref{fig:task} (d1), (d2) and (d3) show the motivation of the proposed method, where the increase of feature dimension and diversity is beneficial for classification robustness against input variations and adversarial attacks.
To increase feature representation space, group-wise feature regularization is proposed to leverage the independence among group-wise features. To improve feature diversity, the proposed algorithm regularizes group-wise features dynamically in each training step.
Specifically, based on the grouping of  feature maps and their importance evaluation, the group-wise reversed map is proposed to suppress the activation values corresponding to the most significant discriminative regions, and
guide the network to learn more auxiliary information in less significant regions.
As shown in the second row of Fig. \ref{fig:problem}, the suppression of most significant discriminative regions is beneficial for exploring more  diverse features in CNNs.
Experimental results show that the proposed method can improve the top-1 error rate of adversarial training from 36.37\% to 31.75\%, and outperforms regularization methods significantly in terms of classification accuracy based on small sample. In summary,

\begin{itemize}
  \item A group-wise inhibition based regularization method is proposed to explore auxiliary features and promote feature diversity.
  \item Feature maps with different activation distribution are processed separately to learn richer discriminative features hierarchically to better represent images.
  \item Our proposed method achieves competitive performances in terms of adversarial robustness and generalization compared with related variants and the state of the arts.
\end{itemize}

\section{Related Work}
\label{sec:RelatedWork}
\subsection{Robustness against Corruption and Adversarial Attack}
The human vision system is robust in ways that CNN based computer vision systems are not \cite{hendrycks2019benchmarking}.
Particularly, a large mount of studies \cite{dong2020benchmarking,jefferson2020robust,hendrycks2019benchmarking,goodfellow2014explaining} show that CNNs can be easily fooled by small variations in query images, including common corruption \cite{hendrycks2019benchmarking} and adversarial perturbation \cite{goodfellow2014explaining}.
In order to improve the robustness against these variations, studies have been proposed based on various strategies, such as structure modification, adversarial training and regularization.
Xie et al. \cite{xie2019feature} proposed a non-local feature denoising block to suppress the disturbation caused by the malicious perturbation. A Discrete Wavelet Transform (DWT) layer is proposed by Li et al. \cite{li2020wavelet}, which disentangles the low- and high-frequency components to yield the noise-robust classification. Different from structure based methods, adversarial training and regularization methods can improve the robustness without the modification of network structure. Adversarial training proposed by Goodfellow et al. \cite{goodfellow2014explaining}, in which a network is trained on adversarial examples, is reported to be able to withstand strong attacks \cite{shafahi2019adversarial}.
However, there is a trade-off between classification accuracy (generalization) and adversarial robustness.
Hence, more and more studies are resorted to the regularization solutions \cite{zhong2020random,hendrycks2019augmix,devries2017improved,hinton2012improving} to simultaneously improve generalization and robustness against variations, i.e. common corruption and adversarial attack.

\subsection{Regularization for CNNs}
Regularization \cite{zhong2020random,hendrycks2019augmix,devries2017improved,hendrycks2020many,hinton2012improving,Tompson_2015_CVPR,hou2018self,wei2017object} has been widely employed in the training of CNNs, where image-wise and feature-wise regularization methods were proposed to improve generalization or robustness.
Data augmentation is a typical image-wise solution to regularize the data distribution \cite{zhong2020random,hendrycks2019augmix,devries2017improved,hendrycks2020many}.
Devries et al. \cite{devries2017improved} proposed a regularization technique to randomly mask out square regions of input during training. Random Erasing proposed by Zhong et al. \cite{zhong2020random} randomizes the values of pixels in a random rectangle region.
Hendrycks et al. \cite{hendrycks2019augmix} proposed Augmix to coordinate simple augmentation operations with a consistency loss.
In a nutshell, these image-wise regularization solutions generate images by random operations (e.g. cutout, erasing and mixing), which concentrate on the global information without fully exploring the independence of local features. Meanwhile, the random operations are not dynamically adapted during the training, which limit the feature diversity. These studies motivate us to enhance the feature diversity to improve network robustness and generalization performances.

To explore local information during regularization, feature-wise regularization techniques, including attention based dropout \cite{choe2019attention}, self-erasing \cite{hou2018self,wei2017object} and group orthogonal training \cite{chen2017training}, are proposed.
Attention based dropout proposed by Choe et al. \cite{choe2019attention} utilizes the self-attention mechanism to regularize the feature maps.
Self-erasing \cite{hou2018self,wei2017object} is an extension method of popular class activation map (CAM) \cite{zhou2016learning,selvaraju2017grad}, which erases the most discriminative part of CAM, and guides the CNNs to learn classification features from auxiliary regions and activations \cite{wang2020self}.
However, these methods are proposed for semantic segmentation rather than the classification task. Meanwhile, the steep gradients introduced by the binary mask limit the performances of dropout and erasing operation for classification task. From another aspect, the erasing operation and dropout are global regularizers, which do not fully explore the independence of feature semantics, i.e. different feature groups contain different semantics and should be processed specifically.
Group orthogonal training proposed by Chen et al. \cite{chen2017training} provides a solution for this problem, which guides CNNs to learn discriminative features from foreground and background separately. Although this group orthogonalization strategy brings improvement of classification performance by enhancing feature diversity, the relied large annotation limits its applicability for general tasks.

In this paper, a regularization method based on group-wise inhibition, namely TENET Training, is proposed to improve network robustness and generalization, which is free of extra annotations. Particularly, a Channel-wise Feature Grouping (CFG) module is proposed to model the channel-wise features in groups. Subsequently, the features in different groups are processed specifically by Group-wise Map Weighting (GMW) module to quantify the importance of each group. Meanwhile, in order to avoid the steep gradients caused by binary mask, a Rectified Reverse Function (RRF) is proposed to smooth group-wise reversed maps. Finally, these reversed maps are used to suppress the activation values to regularize the learned features. Extensive experiments clearly show the significant improvements in terms of robustness and generalization performances.

\section{Proposed Method}
\label{sec:method}
\begin{figure*}
  \centering
  \includegraphics[width=.95\textwidth]{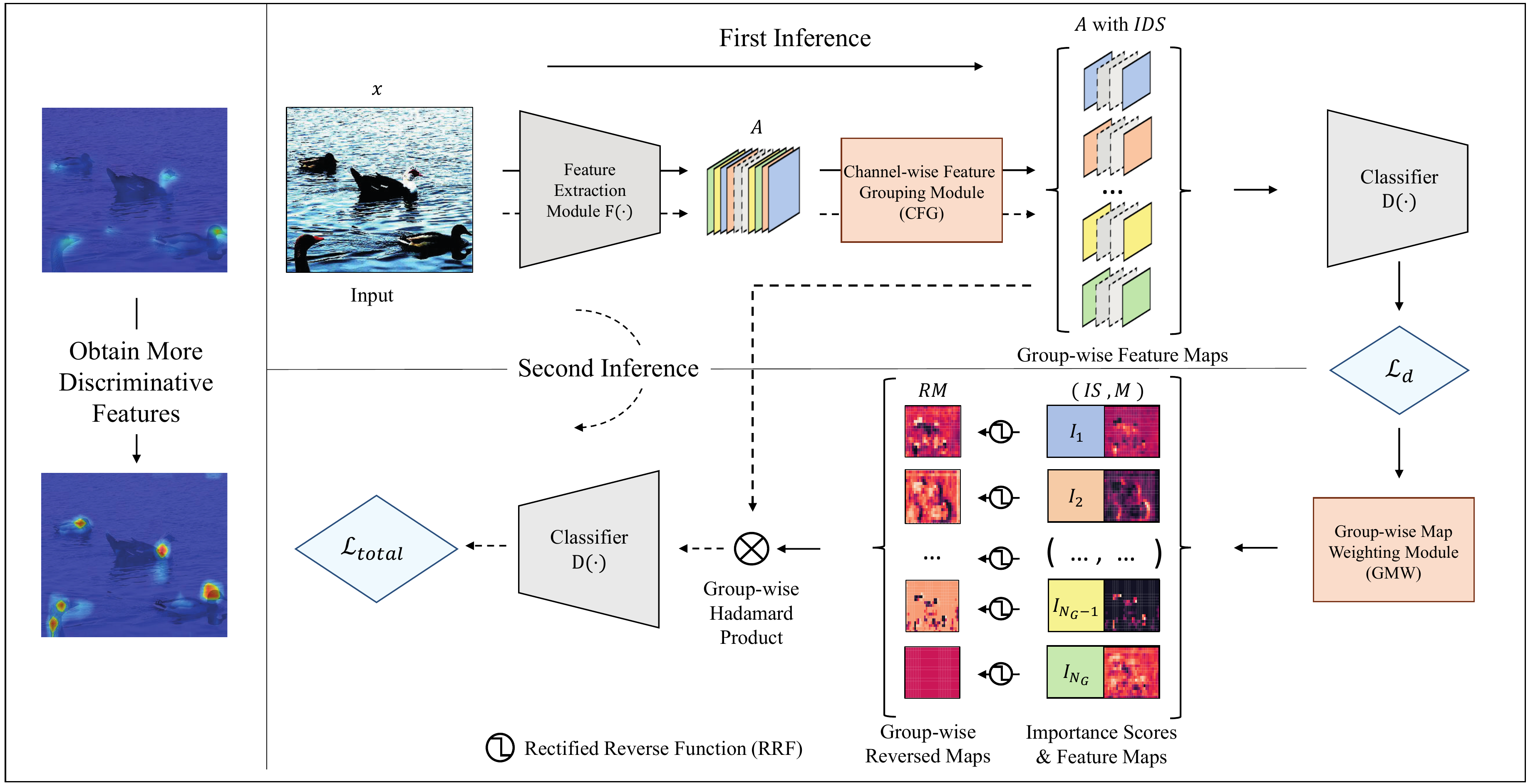}
  \caption{
  The pipeline of the proposed regularization method (TENET Training). Notice that CNNs consist of the feature extraction module $F(\cdot)$ and the classifier $D(\cdot)$. In the first inference, feature maps $A$ encoded with $F(\cdot)$ are divided into $N_G$ groups by the $CFG$ module, and loss $\mathcal{L}_{d}$ is calculated based on $D(\cdot)$. Reversed maps $RM$ are then derived using $GMW$ module and $RRF$. In the second inference, the Hadamard Product of $A$ (with $IDS$) and $RM$ is fed to $D(\cdot)$ to calculate the loss $\mathcal{L}_{total}$.
  }
  \label{fig:pipeline}
\end{figure*}

The overview of the proposed TENET Training is shown in Fig. \ref{fig:pipeline}., where CNN is dynamically regularized according to the training step, and significant activation values are suppressed to guide network to explore different features hierarchically.
Since the feature maps with the similar activation distribution are prone to contain redundant information, we firstly group the channel-wise feature maps using the proposed CFG module in Section \ref{sec:CFG}.
In order to further quantify the contribution of each group, the GMW module is introduced in Section \ref{sec:GMW} to evaluate the group importance.
Considering the feature groups with negative importance score should contribute less to the classification performance, Rectified Reverse Function (RRF) is proposed to smooth the reversed map of the filtered groups. Following RRF, the group-wise inhibition is devised to suppress the most significant features and explores the less significant auxiliary features, which is introduced in Section \ref{sec:RRF}. Finally, we conclude the pipeline of the proposed TENET Training together with the loss design in Section \ref{sec:LD}.

\subsection{Channel-wise Feature Grouping Module}
\label{sec:CFG}
According to the pipeline shown in Fig. \ref{fig:pipeline}, a feature extraction module $F(\cdot)$ is firstly applied to encode the features set $A = \{ a_1, ..., a_j, ... , a_{N_c} \}$ of the input sample $x$, where $a_j$ is the $j$th feature map. Since $A$ is prone to contain redundant features, a Channel-wise Feature Grouping module, denoted as CFG module, is introduced to group $A$ to reduce the complexity of feature-wise operation. Given $N_c$ features as input, the corresponding $N_G$ centers are obtained to form the set $A_c$, which are  initialized as a random subset of $A$. The distance from each feature map of $A$ to the corresponding center is calculated as follows
\begin{align}
  \label{eq:1}
  Dist(a_j,A_c[l]) = \frac{1}{H_a \times W_a} \sum_{H_a} \sum_{W_a} (a_j-A_c[l])^2
\end{align}
where $l \in [1,N_G]$ is the index of the center and ($H_a$, $W_a$) is the size of $a_j$. Based on Eq. (\ref{eq:1}), the centers are updated as similar as k-means clustering. $N_G$ groups are then obtained by grouping the feature maps to the corresponding center. In order to alleviate the influence caused by the random selection, the center searching process is carried out repeatedly in the CFG module. Based on the grouping procedure, the centers are updated according to Center Point Search Function, i.e. CF($\cdot$) as follows
\begin{align}
  \label{eq:2}
  \text{CF}(I\!D\!S) \!=\! \{ \mathop{\arg\min}\limits_{a_j\in A} dist(a_j,\!\frac{1}{n_l}\!\sum_{ID_i=l}\! a_i\! ) \Big| l\! \in\! [1,\!N_G] \}
\end{align}
where the set $IDS = \{ ID_1, ..., ID_j, ..., ID_{N_c}\}$ stands for the set of feature map indices corresponding to each group. $ID_j$ refers to the group index of $a_j$. $n_l$ is the number of feature maps in the $l$th group. Based on Eq. (\ref{eq:2}), $A_c$ can be refined iteratively until CF($\cdot$) is stable.
\subsection{Group-wise Map Weighting Module}
\label{sec:GMW}
Following feature grouping module, the feature maps are processed in the group-wise mode. To differ the contribution of each group, a Group-wise Map Weighting module, namely GMW module, is proposed to calculate the weight $w_j$ of each $a_j$ as follows
\begin{align}
  \label{eq:w}
  \begin{split}
  w_j &= \frac{1}{H_a \times W_a}\sum_{H_a}\sum_{W_a}\frac{\partial \mathcal{L}_d(A) }{\partial a_j} \\
  \mathcal{L}_d(A) &=  D(A) \times \text{One-Hot}(D(A))
  \end{split}
\end{align}
where $D(\cdot)$ is a classifier, which maps $A$ to the class score. $\mathcal{L}_d(A)$ is the product of prediction and the corresponding one-hot vector of $D(A)$. Since $\frac{\partial \mathcal{L}_d(A)}{\partial a_j}$ is applied to quantify the importance of $a_j$ to the prediction, the group-wise importance scores, i.e. $IS = \{ I_1, ..., I_l,..., I_{N_G}\}$ can be obtained by averaging $w_j$ of each group ($ID_j$=$l$) as follows
\begin{align}
  \label{eq:l}
  I_l = \frac{1}{N_l}\sum_{ID_j = l} w_j
\end{align}
Similar to $IS$, the group-wise feature maps, i.e. $M = \{ m_1, ..., m_l, ..., m_{N_G} \}$ can be obtained by averaging the  weighted feature maps as follows
\begin{align}
  \label{eq:m}
  m_l = \frac{1}{N_l}\sum_{ID_j = l} w_j \times a_j
\end{align}
\subsection{Group-wise Inhibition using Rectified Reverse Function}
\label{sec:RRF}
Based on the importance scores, group-wise feature maps are applied to obtain the reversed map set, i.e.  $RM = \{ rm_1, ..., rm_l,...,rm_{N_G} \}$. Since the steep gradients introduced by the binary mask may limit the classification performance, the reversed maps are further smoothed. Meanwhile, considering the feature groups with negative importance scores should contribute less to the update of the reversed mask, we therefore propose a Rectified Reverse Function, i.e. RRF($\cdot$), to obtain the reversed maps as follows
\begin{align}
  \label{eq:rrf}
  rm_l = \text{RRF}(m_l,I_l) = sgn(I_l>0) \times \frac{1}{1+e^{m_l}}
\end{align}
where $sgn(\cdot)$ is the sign function.
Due to the negative correlation between $m_l$ and $rm_l$, the computation of $RM$ is deemed as a reversed map. Based on $RM$, the group-wise inhibition is formulated as follows
\begin{align}
  \label{eq:y}
  \hat{y} = D(RM \otimes A)
\end{align}
where $D(\cdot)$ is a classifier with the input of $A$ and $\hat{y}$ refers to the predicted label of the group-wise inhibition. $\otimes$ refers to the group-wise Hadamard product.
\subsection{Loss Design of TENET Training}
\label{sec:LD}
While $\hat{y}$ is obtained by group-wise inhibition, $F(\cdot)$ and $D(\cdot)$ can be directly learned based on the loss $\mathcal{L}_c(y,\hat{y})$, i.e. the cross entropy for single-label classification or binary cross entropy for multi-label classification. The group-wise inhibition reduces the variation between groups, while it may introduce invalid activation units  in $F(\cdot)$ or $D(\cdot)$. To regularize these activation units, an orthogonal loss $\mathcal{L}_o(A)$ is adopted, which is formulated as follows
\begin{align}
  \label{eq:or}
  \mathcal{L}_o(A) = \prod \limits_{l=1}^{N_g}(\sum_{j=1}^{N_c}(sgn(ID_j = l) \times a_j))
\end{align}
From another aspect, by mapping $rm_l$ into the region of [0, 1], the magnitude of back-propagation gradients is suppressed for $F(\cdot)$ and $D(\cdot)$. To alleviate vanishing gradient problem, a general classification loss, i.e. $\mathcal{L}_c(y_i, D(A))$, is employed. Finally, the total loss is formulated as follows
\begin{align}
  \mathcal{L}_{total} = \mathcal{L}_c(y_i, D(A))+ \alpha \mathcal{L}_c(y_i,\hat{y})  + \mu \mathcal{L}_o(A)
  \label{eq:total}
\end{align}
where $\alpha$ and $\mu$ are the hyper parameters. For clarity, TENET Training is summarized in Algo. 1
\begin{algorithm}[!htb]
  \label{algo:1}
	\caption{TENET Training}
  \begin{algorithmic}
      \Require\\
    Training Sample: $x$ \\
    Initialization of $F(\cdot)$ and $D(\cdot)$
     \Ensure \State Trained CNNs: $F(\cdot)$ and $D(\cdot)$
  \end{algorithmic}
  \begin{algorithmic}[1]
    \For {all training steps}
      \State Extract $A$ from $F(x)$;
      \State Obtain $IDS$ of $A$ using CFG Module according to Eqs. (\ref{eq:1}) and (\ref{eq:2});
      \State Derive $(IS,M)$ with GMW Module according to Eqs. (\ref{eq:w}), (\ref{eq:l}) and (\ref{eq:m});
      \State Employ RRF to obtain $RM$ according to Eq. (\ref{eq:rrf});
      \State Obtain $\hat{y}$ according to Eq. (\ref{eq:y});
      \State Calculate $\mathcal{L}_{total}$ according to Eqs. (\ref{eq:or}) and (\ref{eq:total});
      \State Update $F(\cdot)$ based on  $\frac{\partial \mathcal{L}_{total}}{\partial F}$ and update $D(\cdot)$ based on $\frac{\partial \mathcal{L}_{total}}{\partial D}$;
    \EndFor
      \State Return $F(\cdot)$ and $D(\cdot)$.
	\end{algorithmic}
\end{algorithm}

\section{Experimental Results and Analysis}
\label{sec:experiment}
\begin{table}[!htbp]
  \centering
  \caption{Summary of Experiment Configurations and TENET Training Gains.}
  \Huge
  \resizebox{.48\textwidth}{14.6mm}{
   \begin{threeparttable}
  \begin{tabular}{c|ccc}
  \toprule
 Task-[protocol] & Dataset & Previous SOTA & Gain\\
  \midrule
  \textbf{Standard Classification}-\cite{chen2017training}& \tabincell{c}{PASCAL \\ VOC  2012\cite{everingham2010pascal}}  &  \tabincell{c}{Group Orthogonal \\ Training \cite{chen2017training}}& \textbf{2.9\%} \\
  \midrule
   \textbf{Robustness} against   &  \multirow{2}{*}{CIFAR-10/100 \cite{krizhevsky2009learning}} & A. T. \cite{shafahi2019adversarial}  & \textbf{5.75\%}\\
   Adversarial Attack-\cite{shafahi2019adversarial,dong2020benchmarking} &  & Augmix\cite{hendrycks2019augmix} & \textbf{15.56\%}$^*$\\
  \midrule
  \textbf{Robustness} against & CIFAR-10/100-C \cite{hendrycks2019benchmarking} & \multirow{2}{*}{Augmix\cite{hendrycks2019augmix}} & \textbf{1.77\%}\\
  Common Corruption-\cite{hendrycks2019benchmarking,hendrycks2019augmix,li2020wavelet} &  ImageNet-C \cite{hendrycks2019benchmarking} &  & \textbf{2.8\%}$\dag$\\
  \midrule
  \textbf{Generalization}-\cite{azuri2020learning} & CUB-200 \cite{wah2011caltech}  & GLICO \cite{azuri2020learning} & \textbf{2.75\%}\\
  \bottomrule
  \end{tabular}
     \begin{tablenotes}
     \huge
     \item $^*$ The gain is obtained in CIFAR-10 against FGSM (8/255).
     \item $^\dag$ The gain is obtained by following 90-epoch Protocol \cite{li2020wavelet}.
   \end{tablenotes}
  \end{threeparttable}
  }
  \label{tab:summary}
\end{table}
As listed in Table \ref{tab:summary}, to evaluate the performance of the proposed method, extensive experiments are carried on publicly-available data sets, including PASCAL VOC 2012 \cite{everingham2010pascal}, CIFAR-10/100 \cite{krizhevsky2009learning}, ImageNet-C \cite{hendrycks2019benchmarking} and CUB-200 \cite{wah2011caltech}. We firstly introduce the employed data sets and  the corresponding implementation details. The performance of the proposed method on standard image classification task is evaluated, and the encoded feature maps are visualized for the algorithm analysis. Finally, both the robustness and generalization performances of the proposed method are evaluated based on the comparison with the state-of-the-art methods.
\begin{table*}[!htbp]
  \centering
  \caption{The Ablation Study of the Proposed Method on the Validation Dataset of Pascal VOC 2012 in terms of Average Precision (\%).}
  \Huge
  \resizebox{.99\textwidth}{11.2mm}{
  \begin{tabular}{cccc|cccccccccccccccccccc|c} 
  \toprule
  Baseline&\tabincell{c}{Channel-wise \\ Inhibition} & \tabincell{c}{Group-wise \\ Inhibition} & $L_o$ & areo & bike & bird & boat & bottle & bus & car & cat & chair & cow & table & dog & horse & mbk & prsn & plant & sheep & sofa & train & tv & \textbf{mean}\\
  \midrule
   $\surd$&$\times$ & $\times$ & $\times$ & 94.8 & 83.8 & 91.5 & 79.4 & 56.6 & 88.2 & 78.9 & 90.8 & 64.8 & 61.5 & 57.9 & 90.9 & 73.7 & 83.8 & 96.0 & 51.6 & 77.1 & 58.2 & 89.8 & 77.1 & 77.1\\
   $\surd$&$\surd$ & $\times$ & $\times$  & 94.2 & 82.8 & \textbf{92.9} & 83.3 & 62.2 & 90.8 & 81.0 & \textbf{92.8} & 71.1 & 74.1 & 63.0 & 88.2 & 83.9 & 88.5 & 93.5 & \textbf{58.4} & \textbf{85.2} & 64.7 & 93.1 & 80.6 & 81.2\\
   $\surd$&$\times$ & $\surd$ & $\times$   & 93.9 & 81.7 & 92.5 & \textbf{83.7} & \textbf{63.8} & 90.9 & 82.7 & 91.5 & 69.5 & 76.4 & 64.6 & 89.6 & \textbf{85.9} & \textbf{89.3} & \textbf{96.5} & 58.1 & 84.6 & 64.5 & 93.2 & \textbf{83.7} & 81.8\\
   $\surd$&$\times$ & $\surd$ & $\surd$    & \textbf{95.6} & \textbf{84.3} & 91.1 & 83.1 & 61.3 & \textbf{91.4} & \textbf{83.2} & 91.6 & \textbf{72.8} & \textbf{77.4} & \textbf{65.9} & \textbf{91.3} & 84.4 & 89.2 & 96.3 & 57.4 & 83.9 & \textbf{67.6} & \textbf{94.5} & 83.1 & \textbf{82.3}\\
  \bottomrule
  \end{tabular}
  }
  \label{tab:ABPASCAL}
\end{table*}
\subsection{Data Sets and Implementation Details}
We evaluate the performance of TENET Training from three aspects, i.e. standard classification, robustness and generalization (see Table \ref{tab:summary}).

\textbf{Standard Classification. }
In this case, ResNet-18 \cite{he2016deep} is selected as the backbone in our TENET Training. PASCAL VOC 2012 \cite{everingham2010pascal}  is used for the evaluation, while 5,717 and 5,823 images are used for the training and validation, respectively. The protocol in \cite{chen2017training} is adopted. The CNNs for evaluation are pretrained on the ImageNet \cite{deng2009imagenet}, and fine-tuned on PASCAL VOC 2012 training set. In the training stage, the shorter side of image is resized to a random value within [256,480] for the scale augmentation. The resized image is then randomly cropped to the size of $224 \times 224$ for the training based on the batch size of 256. In the testing stage, ten-crop testing is used to evaluate the performance.

\textbf{Robustness. }
In this case, the robustness of the proposed algorithm against both adversarial attack and image corruption is evaluated on CIFAR 10/100 \cite{krizhevsky2009learning}, CIFAR 10/100-C \cite{hendrycks2020many} and ImageNet-C \cite{hendrycks2020many}. ResNeXt-29 \cite{xie2017aggregated} and ResNet-50 \cite{he2016deep} are chosen as the backbones.
To test the robustness of the proposed method against adversarial attacks, two popular attacks, FGSM \cite{goodfellow2014explaining} and PGD \cite{athalye2018obfuscated}, are employed. The performance is then evaluated according to the protocol in \cite{dong2020benchmarking}. The perturbation budget ($\epsilon$) is set to $8/255$ or $4/255$ under $l_{\infty }$ norm distance for the two attacks. PGD-K stands for K-step attack with a step size of $2/255$.
Meanwhile, adversarial training is used to defense powerful iterative attacks of PGD. To make the results more convincing, an efficient adversarial training method (free-AT) \cite{shafahi2019adversarial} is adopted, where the \textit{hop step} of free-AT, i.e. $m$, is set to $4$.

Against image corruption, 15 different kinds of corruptions, such as noise, blur, weather and digital corruptions, are performed on CIFAR 10/100-C and ImageNet-C for the evaluation, and each kind of corrupted data has five different severity levels \cite{hendrycks2020many}. We follow the training protocols and evaluation metrics used in Augmix \cite{hendrycks2019augmix} and WResNet50 \cite{li2020wavelet}.
The \textit{Clean Error} is the regular classification error on the original (uncorrupted) test or validation dataset, and
\textit{mCE (Mean Corruption Error)} for CIFAR-10/100-C is the mean over all 15 corruptions. Meanwhile, the \textit{mCE} for ImageNet-C is normalized by the corruption error of AlexNet \cite{krizhevsky2012imagenet}.
Due to the computational efficiency, Augmix without Jensen-Shannon divergence (JSD) loss is implemented.

\textbf{Generalization. }
Since CUB-200 \cite{wah2011caltech} contains only 30 images for each of the 200 species of birds, it is used as a popular benchmark to test the generalization of CNNs.
We follow the protocol in \cite{azuri2020learning}, and evaluate  the generalization with three numbers of samples per class (SPC) for training, i.e. 10, 20 and 30.
For a fair comparison, the same ResNet-50 \cite{he2016deep} in the protocol \cite{azuri2020learning} is adopted as the backbone.
To train the CNNs, the smaller side of the images from CUB-200 is resized to 256, the scaled images are then randomly cropped to the size of $224 \times 224$. In the testing stage, the prediction is based on the center  cropping with the size of $224 \times 224$.
\begin{table*}[!htbp]
  \centering
  \caption{Performance Comparison between the Proposed Method and the State of the Arts on the Validation Dataset of Pascal VOC 2012  in terms of Average Precision (\%).}
  \Huge
  \resizebox{.99\textwidth}{14.0mm}{
  \begin{tabular}{c|cccccccccccccccccccc|c} 
  \toprule
  Model & areo & bike & bird & boat & bottle & bus & car & cat & chair & cow & table & dog & horse & mbk & prsn & plant & sheep & sofa & train & tv & \textbf{mean}\\
  \midrule
  ResNet18\cite{he2016deep} reported in \cite{chen2017training} & 95.2 & 79.3 & 90.2 & 82.8 & 52.6 & 90.9 & 78.5 & 90.2 & 62.3 & 64.9 & 64.5 & 84.2 & 81.1 & 82.0 & 91.4 & 50.0 & 78.0 & 61.1 & 92.7 & 77.5 & 77.5\\
  ResNet18 trained in this paper & 94.8 & 83.8 & 91.5 & 79.4 & 56.6 & 88.2 & 78.9 & 90.8 & 64.8 & 61.5 & 57.9 & 90.9 & 73.7 & 83.8 & 96.0 & 51.6 & 77.1 & 58.2 & 89.8 & 77.1 & 77.1\\
  \midrule
  GoCNN \cite{chen2017training} & \textbf{96.1} & 81.0 & 90.8 & \textbf{85.3} & 56.0 & \textbf{92.8} & 78.9 & 91.5 & 63.6 & 69.7 & 65.1 & 84.8 & 84.0 & 83.9 & 92.3 & 52.0 & 83.9 & 64.2 & 93.8 & 78.6 & 79.4\\
  \midrule
  TENET (Binary Mask) & 93.2 & 83.8 & 91.3 & 83.2 & 59.8 & 91.6 & 79.6 & 90.6 & 66.3 & 75.2 & 62.1 & 89.7 & 84.7 & 88.4 & 96.3 & \textbf{58.0} & \textbf{87.0} & 65.2 & 93.1 & 82.1 & 81.1 \\
  TENET (Instance-wise Inhibition) & 93.1 & 82.7 & \textbf{92.6} & 82.9 & 61.1 & 90.9 & 81.8 & \textbf{91.6} & 70.6 & 73.7 & 63.3 & \textbf{91.5} & \textbf{85.6} & 88.5 & \textbf{96.4} & 56.8 & 85.1 & 61.8 & 93.2 & 82.3 & 81.3 \\
  TENET & 95.6 & \textbf{84.3} & 91.1 & 83.1 & \textbf{61.3} & 91.4 & \textbf{83.2} & \textbf{91.6} & \textbf{72.8} & \textbf{77.4} & \textbf{65.9} & 91.3 & 84.4 & \textbf{89.2} & 96.3 & 57.4 & 83.9 & \textbf{67.6} & \textbf{94.5} & \textbf{83.1} & \textbf{82.3}\\
  \bottomrule
  \end{tabular}
  }
  \label{tab:VOC}
\end{table*}

\textbf{TENET Training.}
For the hyper parameter setting, the cluster number $N_G$ is set to 6, while $\alpha$ and $\mu$ are set as 0.1 and 0.1,  respectively.

The public platform pytorch \cite{paszke2017automatic} is used for the implementation of all the experiments on a  work station with CPU of 2.8GHz, RAM of 512GB and GPU of NVIDIA Tesla V100.

\begin{figure}[!htb]
  \centering
  \includegraphics[width=.38\textwidth]{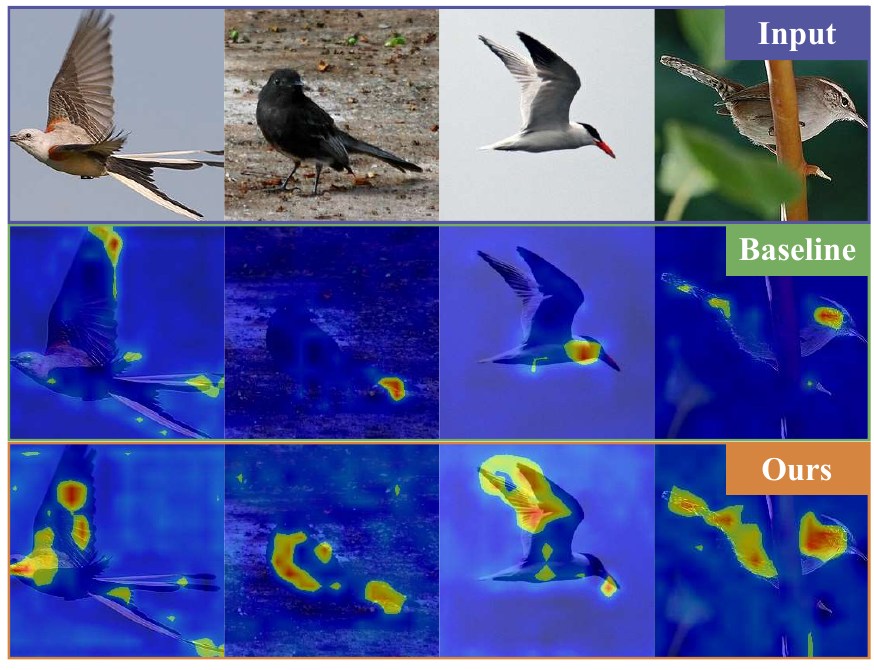}
  \caption{
  The visualization of the discriminative regions for image classification of CUB-200 using Grad-CAM \cite{zhou2016learning,selvaraju2017grad}.
  The 1st-3rd rows show the input samples, the discriminative regions extracted by ResNet-50 and the results based on TENET Training.
  }
  \label{fig:visual_feature}
\end{figure}

\begin{table*}[t]
\centering
  \caption{Top-1 error rates (\%) on ImageNet and Top-1 mCE rates (\%) on ImageNet-C with ResNet-50. Aug. stands for Augmix.}
  \begin{threeparttable}
\Huge
\resizebox{.98 \textwidth}{17.6mm}{
\begin{tabular}{c|c|c|ccc|cccc|cccc|cccc|c}
\hline
\multicolumn{1}{l|}{} & \multirow{2}{*}{Protocol} & Clean         & \multicolumn{3}{c|}{Noise}              & \multicolumn{4}{c|}{Blur}                             & \multicolumn{4}{c|}{Weather}                          & \multicolumn{4}{c|}{Digital}                          & \multicolumn{1}{l}{} \\ \cline{4-19}
\multicolumn{1}{l|}{} &    & Error         & Gauss.      & Shot        & Impulse     & Defocus     & Glass       & Motion      & Zoom        & Snow        & Frost       & Fog         & Bright      & Contrast    & Elastic     & Pixel       & JPEG        & \textbf{mCE}         \\ \hline
\multicolumn{2}{c|}{\tabincell{c}{Baseline \cite{he2016deep} reported in \cite{hendrycks2019augmix}}}       & 23.8          & 79          & 80          & 82          & 82          & 90          & 84          & 80          & 86          & 81          & 75          & 65          & 79          & 91          & 77          & 80          & 80.6                 \\ \hline
Cutout \cite{devries2017improved}                & \multirow{5}{*}{\tabincell{c}{90-epoch \\ Protocol\cite{li2020wavelet}}}       & 23.2          & 79          & 81          & 80          & 77          & 90          & 80          & 81          & 80          & 78          & 70          & 61          & 74          & 87          & 74          & 75          & 77.7                 \\
WResNet50 (Haar) \cite{li2020wavelet}      &        & 23.1          & 77          & 79          & 79          & 71          & 86          & 77          & 77          & 80          & 75          & 66          & 57          & 71          & 84          & 75          & 77          & 75.3                 \\
Augmix \cite{hendrycks2019augmix}               &        & 23.0          & 71          & 71          & 71          & 72          & 88          & 72          & 72          & 78          & 78          & 67          & 60          & 72          & 86          & 75          & 76          & 73.9                 \\
TENET                 &        & 23.1          & 73          & 78          & 75          & 74          & 87          & 76          & 80          & 79          & 78          & 67          & 63          & 73          & 84          & 72          & 71          & 75.3                 \\
TENET (Aug.)          &        & 22.8          & 69          & 69          & 69          & \textbf{69}          & 87          & 69          & 70          & 76          & 75          & \textbf{64}          & \textbf{56}          & 69          & 82          & 72          & 73          & 71.1                 \\ \hline
Augmix \cite{hendrycks2019augmix}                & \multirow{2}{*}{\tabincell{c}{180-epoch \\ Protocol\cite{hendrycks2019augmix}}}       & 22.5          & \textbf{68} & 69          & 70          & 73          & \textbf{81} & 69          & \textbf{67} & 75          & \textbf{73} & 67          & 61          & 61          & \textbf{80} & 71          & 72          & 70.5                 \\
TENET (Aug.)          &       & \textbf{22.4} & 69          & \textbf{67} & \textbf{68} & 72 & \textbf{81} & \textbf{66} & 69          & \textbf{74} & 74          & 65 & 59 & \textbf{60} & 82          & \textbf{69} & \textbf{70} & \textbf{69.6}        \\ \hline
\end{tabular}
}

\end{threeparttable}
\label{tab:image}
\end{table*}

\subsection{Effectiveness Analysis of the Proposed Method}
\begin{figure}[!htb]
  \centering
  \includegraphics[width=.47\textwidth]{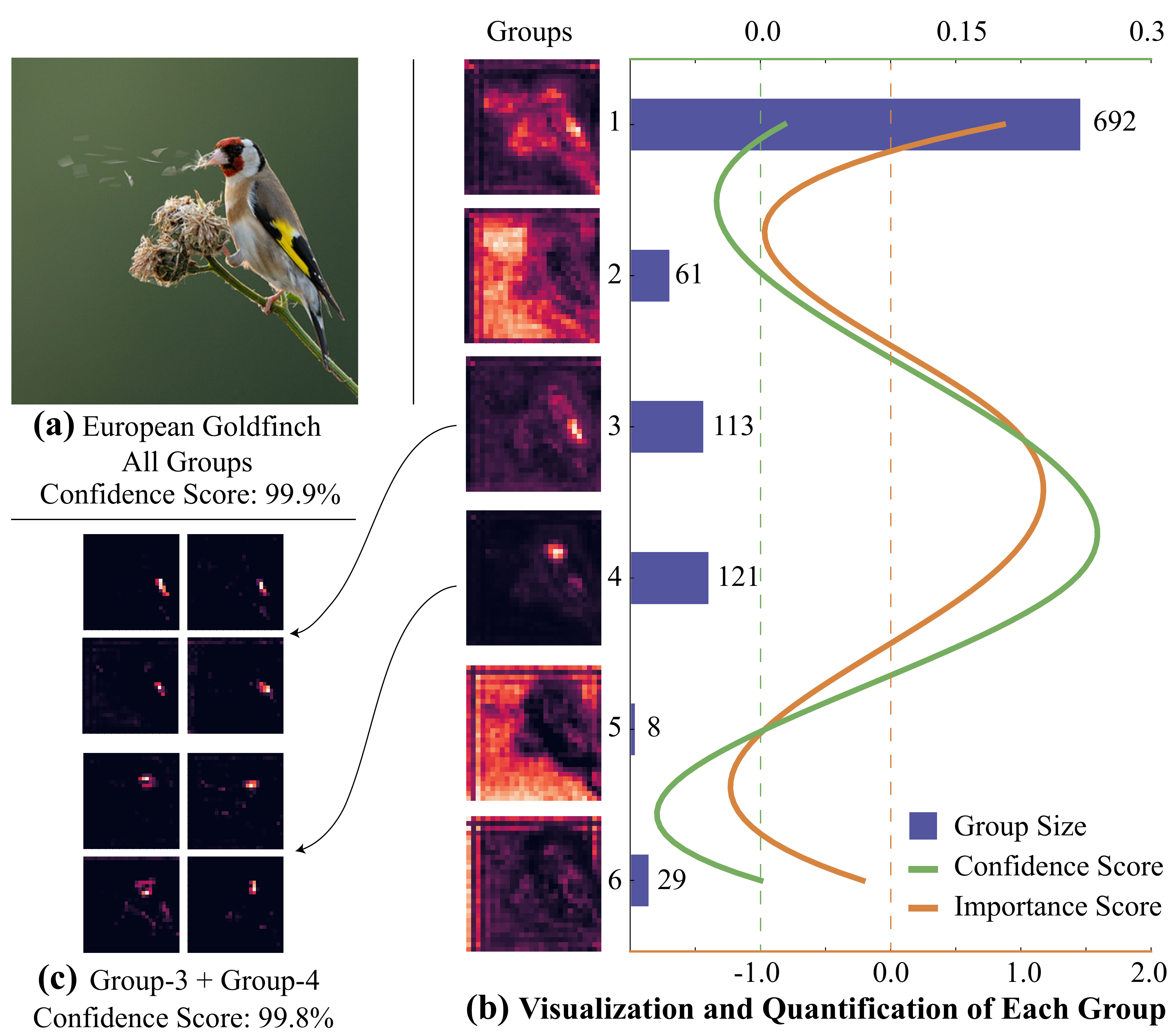}
  \caption{
  The visualization and quantification of the feature maps extracted by the 3rd residual block of ResNet-50 using TENET Training.
  (a) An input image with the label of European Goldfinch. (b) The activation distribution, the corresponding importance and confidence scores of each group clustered by CFG module.
  (c) The example feature maps selected from the 3rd and 4th groups.
  }
  \label{fig:visual_group}
\end{figure}
\textbf{Ablation Study.}
To quantify the contribution of each module in TENET Training, we test the discriminative performance of the variant with or without this module. Table \ref{tab:ABPASCAL} shows the results carried out for standard classification. Since GMW is based on CFG module, these two modules denoted as Channel-wise Inhibition and Group-wise Inhibition are evaluated integratedly.
Table \ref{tab:ABPASCAL} shows
that the performance of the baseline in the first row can be improved by both channel-wise inhibition and group-wise inhibition.
Specifically, an improvement of 4.1\% in terms of mAP is achieved by channel-wise inhibition. To study the performance of GMW and CFG modules, Table \ref{tab:ABPASCAL} shows that the group-wise inhibition further improves the performance using $\mathcal{L}_o(A)$. The most significant improvement of TENET Training happens when all the proposed modules are employed, i.e. the proposed method achieves a mAP of 82.3\%, which largely outperforms the baseline with a mAP of 77.1\%.

\textbf{Visualization of TENET Training.}
To study the diversity of the learned features with the proposed TENET Training, we visualize the discriminative regions of the input samples from CUB-200 using Grad-CAM \cite{zhou2016learning,selvaraju2017grad} in Fig. \ref{fig:visual_feature}. Compared with the baseline, the  CNN using TENET Training derives more discriminative regions, such as wings, heads and tails, for classification.

To study the distribution of the extracted features, we further visualize the group-wise maps with different importance scores of the input image in Fig. \ref{fig:visual_group}, where feature maps are clustered into six groups.
The confidence score of each group corresponds to the variant with or without the selected group. Fig. \ref{fig:visual_group} (b) shows that, the importance score (orange line) calculated by GMW module is similar with confidence score (green line) in tendency, which illustrates the effectiveness of the GMW module.
Meanwhile, Fig. \ref{fig:visual_group} (b) also shows the large variations among the activation distributions of the group-wise features, which indicates the reasonability of group-independent processing. As a contrast, the instance-wise operation involved in traditional methods can not regularize the most important features but only the features with the largest group size ( i.e. group-1 in Fig. \ref{fig:visual_group}), based on the average of activation maps or annotations. Thus, the proposed group-independent processing can facilitate our TENET training to achieve better performance than other regularization methods.

Fig. \ref{fig:visual_group} (c) shows that group-3 and group-4 out of six groups are the most important for CNN, which can improve the confidence score output by the  CNN from 0 to 99.8\%.
Group-1 is relatively less important than group-3 and group-4 but can increase confidence score, while the impacts of group-2, 5 and 6 on the classification performance is very limited. More precisely, when these three groups are not used, the confidence score has dropped by only 0.09\%.
This observation indicates that the inhibition of the important groups can help improve the efficiency without losing accuracy.
Hence, in the proposed method, we only regularize the groups with higher importance scores.


\subsection{Comparison with Related Methods}
\textbf{Comparison in Standard Classification.}
To study the classification performance of the proposed method, we compare it with the group orthogonal training \cite{chen2017training} denoted as GoCNN in Table \ref{tab:VOC}. In additional, we include TENET (Binary Mask) and TENET (Instance-wise Inhibition) for the comparison. TENET (Binary Mask) refers to the proposed method that suppresses the activation value using binary masks rather than the smoothed reversed maps. In TENET (Instance-wise Inhibition), CFG module and GMW module are replaced by Grad-CAM \cite{zhou2016learning,selvaraju2017grad}, which process features by instance-wise operation.
Table \ref{tab:VOC} shows that TENET Training outperforms the competing methods significantly. The proposed method achieves a  mAP of 82.3\%, exceeding group orthogonal training by 2.9\% absolutely. This indicates group-wise inhibition using the smoothed reversed maps is suitable for classification.
Meanwhile, the proposed method uses less information than group orthogonal training, i.e. large-scale dense annotations, e.g. segmentation or localization labels, are not demanded.
While state-agnostic inhibition used in group orthogonal training regularizes features in a coarse way,
it limits both the accuracy and efficiency.
However, based on the proposed group-wise inhibition, our method can consistently improve the classification performance, and does not demand any extra annotations.

\begin{table}[!htbp]
\centering
  \caption{Top-1 error rates (\%) on CIFAR-10 and Top-1 mCE rates (\%) on CIFAR-10-C trained with various methods based on ResNeXt-29. A.T. stands for Adversarial Training. The brackets following the adversarial attack method show the perturbation budget ($\epsilon$). }
\resizebox{.45\textwidth}{23.1mm}{
\begin{tabular}{c|c|c|ccc}
\toprule
                                                         & Clean & mCE  & \begin{tabular}[c]{@{}c@{}}FGSM\\ (8/255)\end{tabular} & \begin{tabular}[c]{@{}c@{}}PGD-7\\ (4/255)\end{tabular} & \begin{tabular}[c]{@{}c@{}}PGD-100\\ (8/255)\end{tabular} \\
\midrule
Baseline \cite{xie2017aggregated}                        & 5.72  & 29.88 & 72.81 & 94.15 & -     \\
Cutout \cite{devries2017improved}                        & 3.97  & 29.20 & 71.07 & 97.19 & -     \\
Augmix \cite{hendrycks2019augmix}                        & 3.95  & 13.32 & 76.03 & 93.67 & -     \\
TENET                                                    & 3.89  & 26.46 & 61.05 & 91.28 & -     \\ \hline
\begin{tabular}[c]{@{}c@{}}TENET\\ (Aug.)\end{tabular}   & \textbf{3.50}  & \textbf{12.31} & 60.47 & 90.45 & -     \\
\toprule
A.T. \cite{shafahi2019adversarial}                       & -  & - & 36.37 & 22.61 & 42.82   \\ \hline
\begin{tabular}[c]{@{}c@{}}TENET\\ (A.T.)\end{tabular}   & -  & - & \textbf{31.75} & \textbf{20.07} & \textbf{37.07}   \\
\bottomrule
\end{tabular}
}
\label{tab:Cifar10}
\end{table}

\textbf{Comparison in Robustness.}
We compare the proposed method with two state-of-the-art regularization methods \cite{devries2017improved,hendrycks2019augmix}, a wavelet integrated method \cite{li2020wavelet} and an adversarial training one \cite{shafahi2019adversarial}, for robustness evaluation against image corruption and adversarial attacks in Tables \ref{tab:image}, \ref{tab:Cifar10} and \ref{tab:Cifar100}. One can observe that TENET Training outperforms the competing methods in each case. For the recognition against image corruption, the best performance is achieved with the combination of TENET Training and Augmix (denoted as TENET(Aug.)), which achieves 69.6\%, 12.31\% and 35.73\% error rates on ImageNet-C, CIFAR-10-C and CIFAR-100-C, respectively.
Augmix \cite{hendrycks2019augmix} with JSD loss can achieve a mCE of 68.4\% on ImageNet-C, while it requires three times the GPU memory and runtime cost compared with the proposed method.

For robustness against adversarial attacks, two attack paradigms, namely FGSM and PGD, are employed to test the trained CNNs with different regularization methods. Tables \ref{tab:Cifar10} and \ref{tab:Cifar100} show that the CNNs using the proposed method outperform those with other regularization methods by a large margin. When FGSM is considered, our method can achieve an error rate of 60.47\%, exceeding other regularization methods by around 10\% absolutely. Meanwhile, our method is complementary to the Adversarial Training (denoted as A.T.). Typically, the proposed method achieves the error rates of 37.07\% and  63.13\% against PGD-100 on CIFAR-10/100, which outperforms Adversarial Training clearly, i.e. 37.07\% vs. 42.82\% and 63.13\% vs. 65.17\%.
\begin{table}[!htbp]
\centering
  \caption{Top-1 error rates (\%) on CIFAR-100 and Top-1 mCE rates (\%) on CIFAR-100-C trained with various methods based on ResNeXt-29.}
\resizebox{.45\textwidth}{23.0mm}{
\begin{tabular}{c|c|c|ccc}
\toprule
                                                         & Clean & mCE  & \begin{tabular}[c]{@{}c@{}}FGSM\\ (8/255)\end{tabular} & \begin{tabular}[c]{@{}c@{}}PGD-7\\ (4/255)\end{tabular} & \begin{tabular}[c]{@{}c@{}}PGD-100\\ (8/255)\end{tabular} \\
\midrule
Baseline \cite{xie2017aggregated}                        & 23.33  & 53.40 & 85.93 & 95.96 & -     \\
Cutout \cite{devries2017improved}                        & 20.73  & 54.60 & 87.03 & 98.13 & -     \\
Augmix \cite{hendrycks2019augmix}                        & 21.83  & 37.50 & 84.65 & 95.32 & -     \\
TENET                                                    & 20.56  & 51.21 & 78.71 & 94.62 & -     \\ \hline
\begin{tabular}[c]{@{}c@{}}TENET\\ (Aug.)\end{tabular}   & \textbf{19.46}  & \textbf{35.73} & 75.28 & 93.54 & -     \\
\toprule
A.T. \cite{shafahi2019adversarial}                       & -  & - & 60.13 & 47.99 & 65.17  \\ \hline
\begin{tabular}[c]{@{}c@{}}TENET\\ (A.T.)\end{tabular}   & - & - & \textbf{58.60} & \textbf{46.17} & \textbf{63.13}  \\
\bottomrule
\end{tabular}
}
\label{tab:Cifar100}
\end{table}


\begin{table}[!htbp]
  \centering
  \caption{Comparison of TOP-1 Accuracy (\%) for CUB-200 based on ResNet-50 with Different Numbers of Training Samples Per Class (SPC). }
    \resizebox{.42\textwidth}{18.4mm}{
  \begin{tabular}{cccc} 
  \toprule
  Methods & SPC = 10 & SPC = 20 & SPC = 30 \\
  \midrule
  MixMatch \cite{berthelot2019mixmatch} & 36.02 & 60.57 & 70.41 \\
  Random Erase \cite{zhong2020random} & 63.72 & 66.14 & 73.74 \\
  Cutout \cite{devries2017improved} & 64.33 & 68.47 & 74.97 \\
  \midrule
  GLICO \cite{azuri2020learning} & 65.13 & 74.16 & 77.75 \\
  \midrule
  A.T. \cite{shafahi2019adversarial} & 44.53 & 57.91 & 63.67 \\
  \midrule
  TENET & \textbf{66.07} & \textbf{76.91} & \textbf{80.34}\\
  \bottomrule
  \end{tabular}
  }
  \label{tab:generation}
\end{table}

\textbf{Comparison in Generalization.}
To further study the generalization performance achieved by TENET Training, we compare the proposed method with regularization methods \cite{berthelot2019mixmatch,zhong2020random,devries2017improved}, data augmentation method \cite{azuri2020learning} and adversarial training \cite{shafahi2019adversarial} in Table \ref{tab:generation}.
Table \ref{tab:generation} shows the evident improvements of TENET Training over other methods in every case.
Typically, when 20 samples per class are used for training, the proposed method can achieve 76.91\% in terms of Top-1 accuracy. As a comparison, adversarial training \cite{shafahi2019adversarial} achieves the Top-1 accuracy of only 57.91\% in this case. It seems that adversarial training can improve the robustness, while it may also largely impair the generalization performance.  Hence, Table \ref{tab:generation} illustrates that the proposed method can better maintain the generalization performance compared with other methods.

\section{Conclusion}
\label{sec:conclusion}
In this paper, we proposed a group-wise inhibition based feature regularization method to improve the robustness and generalization of CNNs.
In the proposed algorithm, CNN is regularized dynamically when learning, where the most discriminative regions with significant activation values are suppressed to enable the network to explore more diverse features.
Richer features then help to better represent images even with malicious variations.
The effectiveness of the proposed method was verified in terms of standard classification, adversarial robustness and generalization performance based on small number of training samples.

\section*{Acknowledgment}
The work is partially supported by the National Natural Science Foundation of China under grants no. 62076163, 91959108, 61602315 and U1713214,the Science and Technology Project of Guangdong Province under grant no. 2020A1515010707, the Shenzhen Fundamental Research fund JCYJ20190808163401646, JCYJ20180305125822769 and JCYJ20190808165203670, and Tencent “Rhinoceros Birds”-Scientific Research Foundation for Young Teachers of Shenzhen University.
{\small
\bibliographystyle{ieee_fullname}
\bibliography{ref_TENET}
}

\end{document}